# Graph Convolution Networks Using Message Passing and Multi-Source Similarity Features for Predicting circRNA-Disease Association

Thosini Bamunu Mudiyanselage
*Department of Computer Science*
Georgia State University
Atlanta, USA
tbamunumudiyanselag1@gsu.edu

Xiujuan Lei
*School of Computer Science*
Shaanxi Normal University
Xi'an 710119, China
xjlei@snnu.edu.cn

Nipuna Senanayake
*Department of Computer Science*
Georgia State University
Atlanta, USA
ssenanayake1@student.gsu.edu

Yanqing Zhang
*Department of Computer Science*
Georgia State University
Atlanta, USA
yzhang@gsu.edu

Yi Pan*
*Department of Computer Science*
Georgia State University
Atlanta, USA
yipan@gsu.edu

*Abstract*— Graphs can be used to effectively represent complex data structures. Learning these irregular data in graphs is challenging and still suffers from shallow learning. Applying deep learning on graphs has recently showed good performance in many applications in social analysis, bioinformatics etc. A message passing graph convolution network is such a powerful method which has expressive power to learn graph structures. Meanwhile, circRNA is a type of non-coding RNA which plays a critical role in human diseases. Identifying the associations between circRNAs and diseases is important to diagnosis and treatment of complex diseases. However, there are limited number of known associations between them and conducting biological experiments to identify new associations is time consuming and expensive. As a result, there is a need of building efficient and feasible computation methods to predict potential circRNA-disease associations. In this paper, we propose a novel graph convolution network framework to learn features from a graph built with multi-source similarity information to predict circRNA-disease associations. First we use multi-source information of circRNA similarity, disease and circRNA Gaussian Interaction Profile (GIP) kernel similarity to extract the features using first graph convolution. Then we predict disease associations for each circRNA with second graph convolution. Proposed framework with five-fold cross validation on various experiments shows promising results in predicting circRNA-disease association and outperforms other existing methods.

## I. INTRODUCTION

Non-Euclidian structure of graphs has the capability to express complex data representations in more meaningful ways. But learning problems on graphs are challenging due to the complex and irregular nature of the underlying patterns of these data. One approach is to convert the graph into lower dimensional space and use traditional learning methods on the reduced space. This process can be affected by loss of some graph properties and suffer from shallow learning. On the other hand, deep learning has good performance in many applications such as convolution neural networks which can extract complex and high-level features in data like images. As a result, more researchers are working on studies to integrate deep learning into graph problems. Graph Convolution Neural Network [1, 2] is such a powerful tool which is able to learn graph data with irregularities using various types of convolutional operators. Message passing convolution is one type of convolution method which aggregates neighbor nodes information to learn global structure of the graph eventually.

RNA-Disease network with complex relations is a good candidate for a graph data representation and can be used to predict RNA-disease associations. This is because of the fact that non-coding RNAs that form the majority of human genome transcripts involve in a wide range of biological processes and evidence proves that they are correlated with many complex human diseases. As a result, there is an increasing interest to develop many computational methods to discover the potential associations between diseases and RNAs. Using these computational methods, we can enhance biomarker discovery for various diseases and limit the expensive wet lab experiments only to verify predicted associations using computational methods.

There are three types of non-coding RNAs: small non-coding microRNAs(miRNAs), long non-coding RNAs(lncRNAs) and circular RNA (circRNA). Most of the existing research has focused on miRNA and lncRNA and they can be broadly divided into two categories: known ncRNA-disease association based and machine learning based approaches. The first category of methods is using known ncRNA-disease associations to forecast potential associations. It assumes that similar RNAs or diseases have similar connection patterns. As an example, if disease 1 and RNA 1 are associated and disease 2 and disease 1 are very similar, it is obvious that RNA 1 and disease 2 are also related. Due to the lack of existing known associations, some prediction methods tend to use biological information related to RNAs such as genome location and tissue specificity [3-7]. As an example, the RNAs close to each other in the genome are frequently associated with the same disease. However, these prediction methods need to collect more data to correctly find the similarity. So, this is not applicable to diseases and RNAs in which tissue specific gene records are not available. Also, the performance of the methods leverages on the correctness of the similarity definitions. Another type of method in this category is to build a bio network fusing disease similarity and RNA similarity sub networks and use methods like random walk [8-11]. In addition to the challenges mentioned above, some information can be lost while fusing these sub networks. The second category of methods utilizes machine learning

This work was supported by the Molecular Basis of Disease Fellowship at Georgia State University.(*Corresponding author: Yi Pan)

algorithms to learn the association between RNAs and diseases. Many machine learning algorithms such as naïve base classifier and support vector machine have been applied in this area. However, these methods use only a part of RNA and disease information and do not fuse them completely and the performance can be affected because of the unbalance distribution of data. Anyway, above methods fail to apply to a new disease without any related RNA. Matrix factorization (MF) [12-15] is one of the most popular machine learning based methods. Traditional matrix factorization methods learn only linear features between RNAs and diseases. But real associations between RNAs and diseases are too complicated and MF cannot capture such associations and they do not get the use of integrating biological information from different cores.

circRNA is a new type of endogenous non-coding RNA molecules which has been discovered in recent years [16,17] and can be stably present in various types of cells. Therefore, recent studies have shown that circRNA plays a critical role in human diseases and circRNA is a good biomarker for disease diagnosis and treatments. With the increment of experimentally verified circRNA-disease associations, computational models can be built to find new potential circRNA-disease associations. MRLDC [18] is one of the few existing computational frameworks which recognize circRNA-disease associations assuming that similar functioning circRNAs are associated with similar diseases. Yan et al. [19] introduced DWNN-RLS to identify circRNA-disease associations using Regularized Least Squares. KATZHCDA [20], PWCDA and RWRHCD [21] are other methods proposed using a heterogeneous network built with disease phenotype similarity and circRNA expression profiles. Due to the limited number of verified circRNA-disease associations and noise in shallow learning, above models suffer from many false negatives.

However, all the previously mentioned methods are shallow learning methods and cannot extract deep and complex representations of RNA disease associations. Deep learning methods can make better performance in extracting non-linear and complex feature representations. To overcome the limitations, hybrid computational models with deep learning techniques have been developed to extract non-linear features of RNA and diseases [22-25]. As an example, combined model with Convolutional Neural Network (CNN) is able to detect high level features from the input data and also the noise can be filtered out during the feature extraction process of CNN [26-28]. However, most of the previous computational methods failed to deeply integrate topological information of heterogeneous networks made of RNAs and diseases. Meanwhile, graph learning further can capture both global and local representation of features relevant to diseases and RNAs and more interestingly deep graphs open up the ability to apply deep learning operations in the domain of irregular, varying graph data. As a result, there is an increased attention on using deep graph learning to predict the RNA-disease associations as graph convolutional networks are proven to be an effective and efficient model for information propagation in networks. GCNCDA [29] is a work which utilizes graph convolution to predict circRNA-disease associations. They employ FastGCN to extract features of circRNA-disease associations and final predictions are based on another classifier Forest PA. Other works [30] and [31] use spectral based graph convolutions to predict lncRNA-disease and miRNA-disease associations respectively. Both of the above models use spectral based convolutions which makes eigenfunctions of Laplacian matrix different from one graph to another. Also, all the computations of Laplacian matrix need to be processed in one time which makes computational complexity high for large graphs. On the other hand, the FastGCN algorithm assumes that nodes in the graph are identically distributed which prevents capturing the actual graph structure.

In this paper, we propose a spatial based two layer graph convolution network to predict circRNA-disease associations and its main contributions are summarized below.

- First, integrate multi-source information such as circRNA sequence similarity, disease and circRNA GIP similarity to build a circRNA-disease network.

- Then, first convolution layer learns feature representations of each node adhering topological structure of the built circRNA-disease network.

- Next, second convolution layer predicts the associativity of multiple diseases of each circRNA using learnt latent feature vectors.

Thus, the proposed model overcomes the limitations of shallow learning and spectral convolution, and relies on deep graph convolutions for both extracting meaningful features and predicting the associations. Message passing convolution is used to aggregate neighborhood information and eventually to learn the real global graph structure. The results of extensive experiments show that the proposed model can successfully predict new circRNA-disease associations and has comparative performance compared to other models.

II. METHOD OVERVIEW

This section describes the proposed method with the detailed information regarding different similarity value calculations, known cicRNA-disease association data, building circRNA-disease network and message passing convolution. Mainly the process can be divided into four steps: calculation and integration of multiple sources of similarity information, building a circRNA-disease network, apply first layer graph convolution to extract high level features of the network and predicting the circRNA-disease associations with the second layer of graph convolution. Fig. 1 depicts the overview of the method.

*A. circRNA-disease Associations*

Experimentally verified known circRNA-disease association data are extracted from circR2Disease [32]. There were about 900 circRNA-disease associations verified with evidence. All of these associations are in between more than 700 circRNAs and more than 100 diseases. The adjacency matrix of circRNA-disease association is denoted by $AS$. If $AS(i,j)$ is equal to 1 there is an association between circRNA $i$ and disease $j$. Otherwise $AS(i,j)$ is equal to 0.

*B. Multi-source Similarity Features*

*1) circRNA Sequence Similarity*

circRNA related RNA based sequence data are taken from circBase database[33]. Then Needleman-Wunsch method, a base pairing algorithm is used to calculate circRNA sequence similarity. $CS(i,j)$ represents the similarity value between the circRNA $C_i$ and $C_j$. Each similarity value $CS(i,j)$ is normalized using (1) where $NW(i,j)$ is the Needleman-Wunsch algorithm score between circRNA $i$ and $j$.

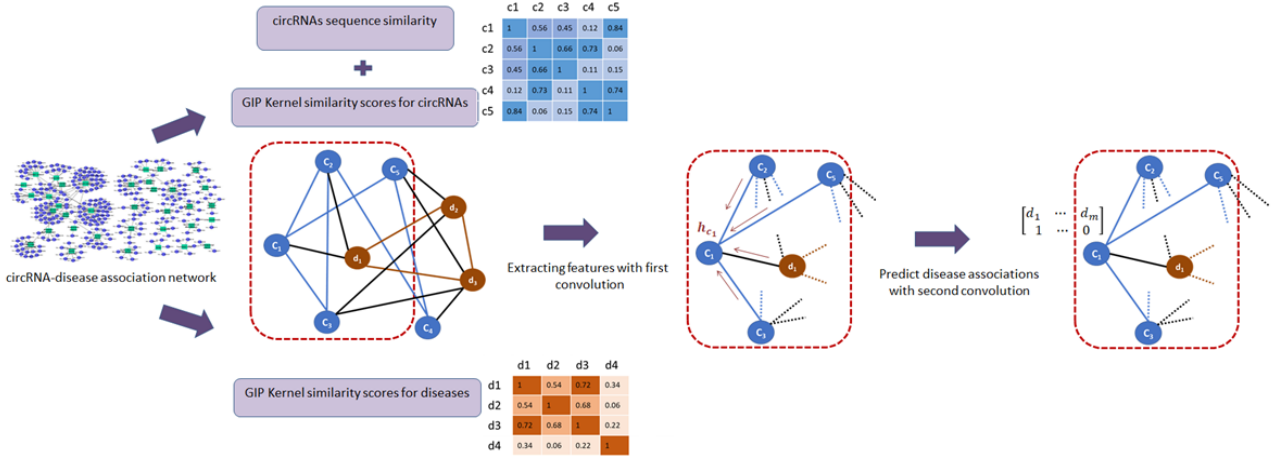

Fig. 1. Overview of the method: 1. Multiple feature similarity calculation and integration, 2 Building circRNA-disease network, 3. Extracting features with first graph convolution, 4. Predict disease associations for each circRNA with second graph convolution.

$$CS(i,j) = \frac{NW(i,j)}{\sqrt{NW(i,i)}\sqrt{NW(j,j)}} \quad (1)$$

*2) circRNA GIP Kernel Similarity*

If two circRNAs are more similar, then they are likely to be associated with the same set of diseases. Based on the above assumption, known circRNA-disease associations are used to calculate circRNA GIP kernel similarity. This similarity score between circRNA $C_i$ and $C_j$ is denoted as $CG(i,j)$ and calculated using matrix $AS$ as mentioned below in (2).

$$CG(i,j) = exp\left(-\alpha_c \left\|P_{C_i} - P_{C_j}\right\|^2\right) \quad (2)$$

Here, $P_{C_i}$ means the interaction profile of circRNA $C_i$ with each disease which is the $i$ th row of matrix $AS$. $\alpha_c$ is an adjusting parameter which controls the range of values and $N_c$ is the total number of circRNAs.

$$\alpha_c = \widehat{\alpha_c} / \left[\frac{1}{N_c}\sum_{i=1}^{N_c}\left\|P_{C_{(i)}}\right\|\right] \quad (3)$$

*3) Disease GIP Kernel Similarity*

GIP kernel similarity algorithm is used to calculate $DG(i,j)$ which is the disease GIP kernel similarity between $D_i$ and $D_j$. The process is similar to calculating circRNA GIP kernel similarity and given in (4).

$$DG(i,j) = exp\left(-\alpha_D \left\|P_{D_i} - P_{D_j}\right\|^2\right) \quad (4)$$

$P_{D_i}$ is the interaction profile of disease $D_i$ and also the $i$ th column vector of matrix $AS$. $\alpha_D$ is the range adjusting parameter and $N_D$ is the total number of diseases.

$$\alpha_D = \widehat{\alpha_D} / \left[\frac{1}{N_D}\sum_{i=1}^{N_D}\left\|P_{D_{(i)}}\right\|\right] \quad (5)$$

*C. Message Passing Graph Convolution*

circRNA-disease network is constructed using calculated similarity features mentioned in above sections. This network is a graph $G = (V, E)$ where $V = v_1, ..., v_n$ which has $n$ number of nodes and $E$ contains $m$ number of edges between nodes considering the calculated similarity feature values. The structural information of this graph can be denoted using an adjacency matrix A. Given a graph $G$, with initial features $X^{n \times d}$ where $d$ is starting dimension of each node $n$, our goal is to map nodes to a lower dimensional feature matrix $Z^{n \times a}$ by learning the topological information of the graph, where $a$ is less than $d$. Message passing convolution operation is used for this purpose and it can be mathematically defined as in (6).

$$h_v^k = f\left(W_k \sum_{u \in N(v), A} h_u^{k-1}\right) \quad (6)$$

Let $h_v^k$ be the output of the $k^{th}$ convolutional layer and input is $h_u^{k-1}$ where $u$ denotes the neighbors of node $v$. Usually, we can start with one-hot feature vectors and use graph structure which consists of different similarities to extract high level features of each node and do the classification based on the calculated features. $W_k$ is the learnable parameters of the $k^{th}$ layer and $f$ is the activation function such as ReLU.

Based on the above facts, we utilize message passing convolution to aggregate different features of neighbors, and eventually learn graph structure to extract high level features of circRNA-disease associations and predict new associations using them. As an example in Fig. 1, node $C_1$ starts with one hot vector and learn the edges with $C_2, C_3, C_5$, $d_1$ and update features of $C_1$ accordingly. This is because we know that $C_1$ is more similar with circRNAs $C_2, C_3, C_5$ and also it is associated with $d_1$ disease based on known associations. Next, first convolution layer converts learned features to lower dimensional high level feature embedding of each node where the second layer gets the output feature embedding of previous layer to do the final predictions. We use already known associations as the output labels to calculate the loss and back propagation to learn the model parameters.

## III. EXPERIMENTS AND RESULTS

We did various experiments on the dataset circR2Disease to evaluate the performance of the graph convolution algorithm in predicting circRNA-disease association. We used five-fold cross validation as it has more reliability in the performance and also avoid overfitting issues. So we divide data into five subsets where each set is taken to test the performance while others are used for training. Thus, we can repeat the process five times and get the average as the model performance. At first, training and testing performance of the model is analyzed and then we did experiments with different number of disease predictions varying from 3 to 50. In the next experiment we show the model performance for each individual disease. Here we selected five diseases having frequent associations with the circRNAs in the dataset to evaluate the model. Hyper-parameter optimization test and comparison with other existing methods are presented in the next section. Finally, we did a case study to further demonstrate the capability of the model on predicting new circRNA-disease association based on the existing associations. Evaluation metrices include accuracy, precision, recall, f1 score which are defined below.

$$Accuracy = \frac{TP + TN}{TP + TN + FP + FN}$$
$$Precision = \frac{TP}{TP + FP}$$
$$Recall = \frac{TP}{TP + FN}$$
$$F1\ score = \frac{2\ TP}{2\ TP + FP + FN}$$

Where $TP, TN, FP, FN$ are true positives, true negatives, false positives and false negatives, respectively. Further we use AUC measure to evaluate the model and compare the performance with others.

### A. Training and Testing Performance

In this experiment we observed model performance in the learning process. Fig. 2(a) depicts the training error against the epoch for different number of disease prediction while Fig. 2(b) plotted both training and testing error on 50 disease predictions with the epoch number. We can observe that training errors of all three models are gradually decreasing as the number of epochs increase. Though 3 disease classification has the highest reduction of training error in early epochs due to its simplicity, all learning curves converge to same error with fewer number of epochs like 20. On 50 disease prediction, both training and testing error converge within 10 epochs and we can observe proposed model has a good learning behavior.

### B. Different Number of Disease Predictions

To evaluate the impact of the increase of number of disease predictions, different models are trained for 3 disease, 10 disease and 50 disease predictions. The performance of each model is measured using 5-fold cross validation and the results are summarized in Tables I, II and III.

In all three different models, average accuracy is more than 96% with very small standard deviation and 98% with three disease prediction. Also, the precision value in most of the test cases and the average precision is very high which denotes that false positives are very less with the graph convolution model in predicting the associativities. Recall values are having lowest performance like 89% score in some test folds and the lowest average is 92% with 50 disease prediction. However, f1 score is a reliable measure which encounters both precision and recall and has stable performance about 95% with the highest standard deviation 0.02. Further capability of the model in predictions can be observed with almost similar AUC 96%.

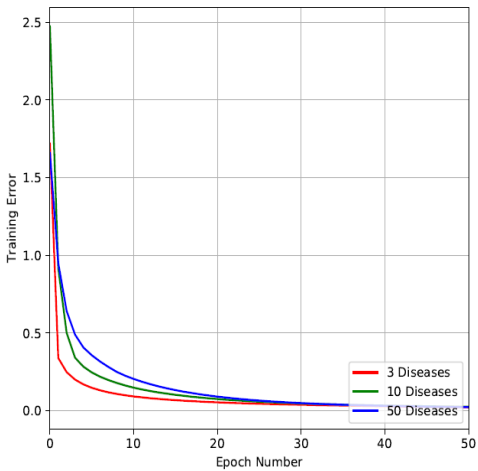

(a) Training error for different number of diseases.

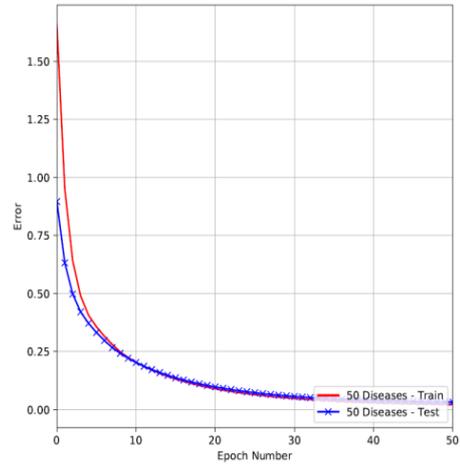

(b) Training and testing error for 50 diseases.

Fig. 2. Performance of the model in the training process

TABLE I. RESULTS OF 5-FOLD CV ON THREE DISEASES PREDICTION

| Test fold | Accuracy | Precision | Recall | F1-score | AUC |
|---|---|---|---|---|---|
| 1 | 0.9877 | 1.0000 | 0.9091 | 0.9524 | 0.9545 |
| 2 | 0.9939 | 1.0000 | 0.9722 | 0.9859 | 0.9861 |
| 3 | 1.0000 | 1.0000 | 1.0000 | 1.0000 | 1.0000 |
| 4 | 0.9877 | 0.9444 | 0.9189 | 0.9315 | 0.9572 |
| 5 | 0.9753 | 0.9048 | 0.9744 | 0.9383 | 0.9827 |
| Average | 0.9889 ± 0.008 | 0.9698 ± 0.039 | 0.9549 ± 0.035 | 0.9616 ± 0.027 | 0.9761 ± 0.018 |

TABLE II. RESULTS OF 5-FOLD CV ON TEN DISEASES PREDICTION

| Test fold | Accuracy | Precision | Recall | F1-score | AUC |
|---|---|---|---|---|---|
| 1 | 0.9877 | 1.0000 | 0.8955 | 0.9449 | 0.9478 |
| 2 | 0.9816 | 1.0000 | 0.9394 | 0.9688 | 0.9697 |
| 3 | 0.9938 | 1.0000 | 0.9870 | 0.9935 | 0.9935 |
| 4 | 0.9877 | 0.9726 | 0.9467 | 0.9595 | 0.9727 |
| 5 | 0.9444 | 0.9487 | 0.9250 | 0.9367 | 0.9612 |
| Average | 0.9791 ± 0.018 | 0.9843 ± 0.021 | 0.9387 ± 0.030 | 0.9607 ± 0.020 | 0.9690 ± 0.015 |

TABLE III. RESULTS OF 5-FOLD CV ON FIFTY DISEASES PREDICTION

| Test fold | Accuracy | Precision | Recall | F1-score | AUC |
|---|---|---|---|---|---|
| 1 | 0.9571 | 1.0000 | 0.8851 | 0.9391 | 0.9426 |
| 2 | 0.9571 | 0.9929 | 0.9329 | 0.9619 | 0.9664 |
| 3 | 0.9753 | 1.0000 | 0.9650 | 0.9822 | 0.9825 |
| 4 | 0.9630 | 0.9790 | 0.9396 | 0.9589 | 0.9696 |
| 5 | 0.9506 | 1.0000 | 0.8929 | 0.9434 | 0.9464 |
| Average | 0.9606 ± 0.008 | 0.9944 ± 0.008 | 0.9231 ± 0.030 | 0.9571 ± 0.015 | 0.9615 ± 0.015 |

Average performance of each metric of above 3 different models is shown in Fig. 3. When the number of disease predictions increased we see slight drop of accuracies but still greater than 95%. Compared to the precision, recall values are little bit lower with all three models while average values of them are greater than 90%. Meanwhile, both f1 score and AUC show stable performance of the model without depending on the number of diseases. Overall, model performs well with all measures above 90%.

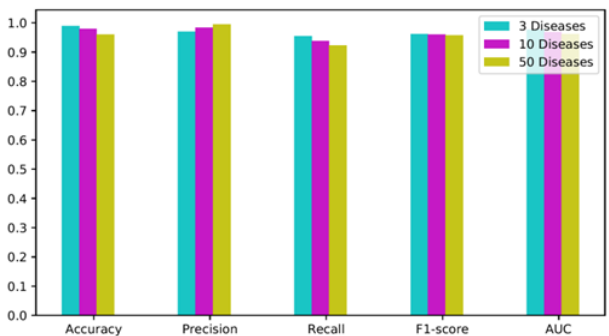

Fig. 3. Comparison of average 5-fold CV results of three, ten and fifty disease predictions

## C. Comparison of Different Disease Prediction Performance

Here we use 50 disease prediction model to compare the performance of individual disease on the test dataset. 5-fold cross validation is utilized with the model and the average performance of each disease is given in Table IV. For the comparison, these values are plotted in a bar chart and shown in Fig. 4.

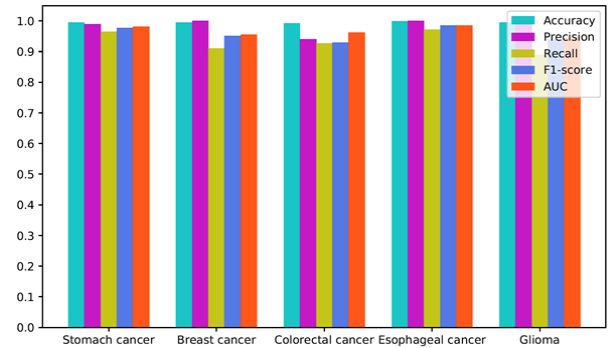

Fig. 4. Comparison of average 5-fold CV results of five individual disease prediction

We see high average accuracy and precision values for all five disease predictions which indicates the effectiveness of the model and low false positives. Both f1 score and AUC are having similar performance except Colorectal Cancer which further proves the robustness of the model. Only Glioma and Colorectal Cancer have lower recall and f1 score with compared to other diseases. But we can conclude that at least three diseases out of five have very high performance while all the measures are greater than or equal to 90% relevant to all diseases.

## D. Comparison with other Methods and Parameter Optimization

TABLE IV.   RESULTS OF 5-FOLD CV ON INDIVIDUAL 5 DISEASES

| Disease | Accuracy | Precision | Recall | F1 | AUC |
|---|---|---|---|---|---|
| Stomach Cancer | 0.9571 | 1.0000 | 0.8851 | 0.9391 | 0.9426 |
| Breast Cancer | 0.9571 | 0.9929 | 0.9329 | 0.9619 | 0.9664 |
| Colorectal Cancer | 0.9753 | 1.0000 | 0.9650 | 0.9822 | 0.9825 |
| Esophageal Cancer | 0.9630 | 0.9790 | 0.9396 | 0.9589 | 0.9696 |
| Glioma | 0.9506 | 1.0000 | 0.8929 | 0.9434 | 0.9464 |

TABLE V.   COMPARISON WITH OTHERS USING 5-FOLD CV GENERATED AUC

| | Our model | GCNCDA | DWNN-RLS | KATZHCDA | PWCDA | RWRHCD |
|---|---|---|---|---|---|---|
| AUC | 0.9615 | 0.9090 | 0.8854 | 0.7936 | 0.8900 | 0.6660 |

Few studies have introduced models with different methods to predict circRNA-disease associations. To compare our model with them, we use 5-fold cv average AUC values on 50 disease prediction model and results are given in Table V.

Based on the results, we can observe that the proposed model effectively predicts potential circRNA related diseases with average 96% AUC. Next, we did hyper-parameter optimization test for the threshold value $\gamma$ which is used to determine the connectivity of two nodes based on the integrated similarity values. If the calculated similarity between nodes $i$ and $j$ is greater than $\gamma$, an edge is built between nodes $i$ and $j$. We used different $\gamma$ values and the average accuracy and f1 score are recorded in the Table VI.

TABLE VI.   COMPARISON OF THE MODEL WITH DIFFERENT $\gamma$ VALUES

| $\gamma$ | 0.01 | 0.1 | 0.2 | 0.5 | 0.8 | 0.9 |
|---|---|---|---|---|---|---|
| Accuracy | 0.8997 | 0.9059 | 0.9359 | 0.9653 | 0.9506 | 0.9444 |
| F1 score | 0.8636 | 0.8857 | 0.8957 | 0.9655 | 0.9306 | 0.9090 |

Above results show that both accuracy and f1 score are having values less than or equal to 90% for $\gamma$ values less than 0.1. When we increase $\gamma$ from 0.1 to 0.5 there is an increment for both values and reaches the highest with 0.5. Again we see a slight drop of accuracy with 0.8 and 0.9. This means too sparse or too dense graphs can degrade the performance of convolution model. Right connectivities with appropriate threshold makes learning better in both feature extraction and prediction. So we selected $\gamma$ as 0.5 in our experiments.

*E. Case Study*

The purpose of this study is to evaluate the capabilities of the model for identifying new circRNA-disease associations based on the existing associations on the training data. Here we select Stomach Cancer for this case study where we calculate prediction probabilities of each association with this disease and sort them in the descending order. Next, we rank each association and top 20 predictions are given in Table VII.

With our computational method, we can see that 16 out of 20 predictions with highest probabilities are already confirmed with evidence from PubMed ID (PMID). For example, CACNA2D1 is proved to be associated with stomach cancer and FGGY is known to down regulate in stomach cancer tissue in [34] and [35], respectively. Further there are four circRNAs are predicted as associated to stomach cancer with high probabilities though they are not yet confirmed.

TABLE VII.   TOP 20 RESULTS FOR STOMACH CANCER BASED ON PREDICTION SCORE

| Rank | circRNA | PMID | Rank | circRNA | PMID |
|---|---|---|---|---|---|
| 1 | CACNA2D1 | 28618205 | 11 | RHBDD1 | 29098316 |
| 2 | FGGY | 28656881 | 12 | TATDN3 | 28940688 |
| 3 | hsa_circ_0000190 | 28130019 | 13 | VIL1 | 28639908 |
| 4 | hsa_circ_0003159 | 28618205 | 14 | ASXL1 | 29109417 |
| 5 | hsa_circ_0005529 | 28737829 | 15 | hsa_circ_0001649 | 28167847 |
| 6 | hsa_circ_0006022 | 28639908 | 16 | CDR1-AS | 28608528 |
| 7 | hsa_circ_0032821 | 28737829 | 17 | hsa_circ_0002113 | - |
| 8 | hsa_circ_0085553 | 28639908 | 18 | circRNA-chr19 | - |
| 9 | hsa_circ_0089548 | 28639908 | 19 | circular_RNA_100783 | - |
| 10 | hsa_circ_001066 | 28184940 | 20 | STAT3 | - |

## IV. CONCLUSION

In this work, we proposed a novel model which uses message passing graph convolutional networks to predict circRNA-disease associations. First, our model gets the advantage of multiple sources of similarity features such as disease and circRNA GIP kernel similarity, circRNA sequence similarity to build a meaningful graph structure and then graph convolution is utilized to extract high level features for each node and finally predict diseases for each circRNA. Experiments on different number of disease predictions and the performance on individual disease show that our model can effectively identify potential associations based on previously known associations with over 90% score for all the metrices: accuracy, precision, recall, f1 score and AUC using 5-fold cross validation. Also, our model has competitive performance with 96% AUC compared to other existing methods and with the case study we can observe that 16 out of 20 highest probability predicted circRNA and stomach cancer associations are confirmed by published literature evidence. Therefore, the proposed computational model can provide potential circRNA-disease associativity effectively based on rich number of existing associations.

## REFERENCES

[1] Z. Wu, S. Pan, F. Chen, G. Long, C. Zhang, and S. Y. Philip. "A comprehensive survey on graph neural networks." IEEE Transactions on Neural Networks and Learning Systems (2020).

[2] T. N. Kipf and M. Welling. "Semi-supervised classification with graph convolutional networks." arXiv preprint arXiv:1609.02907 (2016).

[3] J.W. Li, C. Gao, Y.C. Wang, W. Ma, J. Tu, J.P. Wang, Z.Z. Chen, W. Kong, Q.H. Cui, "A bioinformatics method for predicting long noncoding RNAs associated with vascular disease." Sci. China Life Sci. (2014), 57, 852–857.

[4] A.K. Biswas, B. Zhang, X. Wu, J.X. Gao, "A multi-label classification framework to predict disease associations of long non-coding RNAs (lncRNAs)." In Proceedings of the Third International Conferenceon Communications, Signal Processing, and Systems, Hohot, China, Springer: Basel, Switzerland, (2015), pp. 821–830.

[5] M.-X. Liu, X. Chen, G. Chen, Q.-H. Cui, G.-Y. Yan, "A computational framework to infer human disease-associated long noncoding RNAs", PloS one, 9 (2014).

[6] W. Lan, J. Wang, M. Li, J. Liu, F. Wu, and Y. Pan. "Predicting microRNA-disease associations based on improved microRNA and disease similarities." IEEE/ACM transactions on computational biology and bioinformatics 15, no. 6 (2016): 1774-1782.

[7] W, Lan, J. Wang, M. Li, J. Liu, Y. Li, F. Wu, and Y. Pan. "Predicting drug–target interaction using positive-unlabeled learning." Neurocomputing 206 (2016): 50-57.

[8] Q. Yao, L. Wu, J. Li, L. guang Yang, Y. Sun, Z. Li, S. He, F. Feng, H. Li, Y. Li, Global prioritizing disease candidate lncRNAs via a multi-level composite network, Scientific reports, 7 (2017) 39516.

[9] X. Chen, KATZLDA: KATZ measure for the lncRNA-disease association prediction, Scientific reports, 5 (2015) 16840.

[10] M. Zhou, X. Wang, J. Li, D. Hao, Z. Wang, H. Shi, L. Han, H. Zhou, J. Sun, Prioritizing candidate disease-related long non-coding RNAs by walking on the heterogeneous lncRNA and disease network, Molecular bioSystems, 11 (2015) 760-769.

[11] M. Li, Y. Lu, J. Wang, F. Wu, and Y. Pan. "A topology potential-based method for identifying essential proteins from PPI networks." IEEE/ACM transactions on computational biology and bioinformatics 12, no. 2 (2014): 372-383.

[12] X. Chen, G.-Y. Yan, Novel human lncRNA–disease association inference based on lncRNA expression profiles, Bioinformatics, 29 (2013) 2617-2624.

[13] G. Fu, J. Wang, C. Domeniconi, G. Yu, Matrix factorization-based data fusion for the prediction of lncRNA–disease associations, Bioinformatics, 34 (2017) 1529-1537.

[14] C. Lu, M. Yang, F. Luo, F.-X. Wu, M. Li, Y. Pan, Y. Li, J. Wang, Prediction of lncRNA-disease associations based on inductive matrix completion, Bioinformatics, 1 (2018) 8.

[15] C. Lu, M. Yang, F. Luo, F. Wu, M. Li, Y. Pan, Y. Li, and J. Wang. "Prediction of lncRNA–disease associations based on inductive matrix completion." Bioinformatics 34, no. 19 (2018): 3357-3364.

[16] X. Fan, X. Weng, Y. Zhao, W. Chen, T. Gan, and D. Xu. "Circular RNAs in cardiovascular disease: an overview." BioMed research international (2017).

[17] J. Greene, A. Baird, L. Brady, M. Lim, S. G. Gray, R. McDermott, and S. P. Finn. "Circular RNAs: biogenesis, function and role in human diseases." Frontiers in molecular biosciences 4 (2017): 38.

[18] Q. Xiao, J. Luo, J. Dai "Computational Prediction of Human Disease-associated circRNAs based on Manifold Regularization Learning Framework. " IEEE Journal of Biomedical and Health Informatics (2019) PP: 1-1.

[19] C. Yan, J. Wang, F-X Wu "DWNN-RLS: regularized least squares method for predicting circRNA-disease associations. " BMC bioinformatics 19: (2018) 520.

[20] C. Fan, X. Lei, F-X Wu "Prediction of CircRNA-Disease Associations Using KATZ Model Based on Heterogeneous Networks. International journal of biological sciences 14: (2018) 1950.

[21] X. Lei, Z. Fang, L. Chen, and F. Wu. "PWCDA: path weighted method for predicting circRNA-disease associations." International journal of molecular sciences 19, no. 11 (2018): 3410.

[22] M. Zeng, L. Chengqian, Z. Fuhao, L. Yiming, W. Fang-Xiang, L. Yaohang, and L. Min, "SDLDA: lncRNA–disease association prediction based on singular value decomposition and deep learning." Methods (2020)

[23] J. Hu, Y. Gao, J. Li and X. Shang "Deep learning enables accurate prediction of interplay between lncRNA and disease." Frontiers in genetics 10 (2019): 937

[24] M. Zeng, C. Lu, Z. Fei, F. Wu, Y. Li, J. Wang, and M. Li. "DMFLDA: A deep learning framework for predicting lncRNA–disease associations." IEEE/ACM Transactions on Computational Biology and Bioinformatics (2020).

[25] C. Fan, X. Lei, Y. Pan. "Prioritizing CircRNA-disease Associations with Convolutional Neural Network Based on Multiple Similarity Feature Fusion", Frontiers in Genetics (2020): doi: 10.3389/fgene.2020.540751, in press

[26] H. Wei, Q. Liao, and B. Liu. "iLncRNAdis-FB: identify lncRNA-disease associations by fusing biological feature blocks through deep neural network." IEEE/ACM Transactions on Computational Biology and Bioinformatics (2020).

[27] P. Xuan, Y. Cao, T. Zhang, R.Kong, and Z. Zhang. "Dual convolutional neural networks with attention mechanisms based method for predicting disease-related lncRNA genes." Frontiers in genetics 10 (2019): 416.

[28] P. Xuan, L. Jia, T. Zhang, N. Sheng, X. Li, and J. Li. "LDAPred: a method based on information flow propagation and a convolutional neural network for the prediction of disease-associated lncRNAs." International journal of molecular sciences 20, no. 18 (2019): 4458.

[29] L. Wang, Z. You, Y. Li, K. Zheng, and Y. Huang. "GCNCDA: A new method for predicting circRNA-disease associations based on Graph Convolutional Network Algorithm." PLOS Computational Biology 16, no. 5 (2020): e1007568.

[30] J. Zhang, X. Hu, Z. Jiang, B. Song, W. Quan, and Z. Chen. "Predicting Disease-related RNA Associations based on Graph Convolutional Attention Network." In 2019 IEEE International Conference on Bioinformatics and Biomedicine (BIBM), (2019), pp. 177-182. IEEE

[31] Y. Ding, L. Tian , X. Lei, et al. Variational graph auto-encoders for miRNA-disease association prediction, Methods (2020);online doi.org/10.1016/j.ymeth.2020.1008.1004, in press

[32] C. Fan, X. Lei, Z. Fang, et al. CircR2Disease: a manually curated database for experimentally supported circular RNAs associated with various diseases, Database the Journal of Biological Databases & Curation (2018) ;2018:bay044.

[33] P. Glažar, P. Papavasileiou, N. Rajewsky "circBase: a database for circular RNAs." Rna 20: (2014) 1666-1670.

[34] M. Tian, R. Chen, T. Li, B. Xiao. "Reduced expression of circRNA hsa_circ_0003159 in gastric cancer and its clinical significance. " J Clin Lab Anal. (2018);32(3):e22281. doi:10.1002/jcla.22281

[35] R. Lu, Y. Shao , G. Ye, B. Xiao, J. Guo. "Low expression of hsa_circ_0006633 in human gastric cancer and its clinical significances." Tumour Biol. (2017);39(6): doi:10.1177/1010428317704175